\documentclass[10pt,a4paper]{article}
\usepackage{subfig}
\usepackage{amsmath}
\usepackage{booktabs}
\usepackage[pdftex]{graphicx}
\usepackage{epstopdf}
\usepackage{enumitem}
\usepackage{hyperref}

\begin{document}

\title{Coverage-Driven Verification ---\\ An approach to verify code for robots that directly interact with humans} 

\author{Dejanira Araiza-Illan\footnote{Department of Computer Science and Bristol Robotics Laboratory, University of Bristol, Bristol, UK. Email:  \href{mailto:dejanira.araizaillan@bristol.ac.uk}{dejanira.araizaillan@bristol.ac.uk}.}, David Western\footnote{Department of Computer Science and Bristol Robotics Laboratory, University of Bristol, Bristol, UK. Email:  \href{mailto:david.western@bristol.ac.uk}{david.western@bristol.ac.uk}}, Anthony Pipe\footnote{Faculty of Engineering Technology and Bristol Robotics Laboratory, University of the West of England, Bristol, UK. Email: \href{mailto:tony.pipe@brl.ac.uk}{tony.pipe@brl.ac.uk}.} and Kerstin Eder\footnote{Department of Computer Science and Bristol Robotics Laboratory, University of Bristol, Bristol, UK. Email:  \href{mailto:kerstin.eder}{kerstin.eder@bristol.ac.uk}.}}

\date{14 September 2015}

\maketitle

\begin{abstract}
Collaborative robots could transform several industries, such as manufacturing and healthcare, but they present a significant challenge to verification. The complex nature of their working environment necessitates testing in realistic detail under a broad range of circumstances.  We propose the use of Coverage-Driven Verification (CDV) to meet this challenge. By automating the simulation-based testing process as far as possible, CDV provides an efficient route to coverage closure. We discuss the need, practical considerations, and potential benefits of transferring this approach from microelectronic design verification to the field of human-robot interaction. We demonstrate the validity and feasibility of the proposed approach by constructing a custom CDV testbench and applying it to the verification of an object handover task.
\end{abstract}

\section{Introduction}\label{sc:introduction}
Human-Robot Interaction (HRI) is a rapidly advancing sector within the field of robotics. Robotic assistants that engage in collaborative physical tasks with humans are increasingly being developed for use in industrial and domestic settings. However, for these technologies to translate into commercially viable products, they must be demonstrably safe and functionally sound, and they must be deemed trustworthy by their intended users~\cite{ROMAN14}. In existing industrial robotics, safety is achieved predominantly by physical separation or through limiting the robot's physical capabilities (e.g., speed, force) to thresholds, according to predefined interaction zones. To fully realize the potential of collaborative robots, the correctness of the software with respect to safety and functional (liveness) requirements needs to be verified. 

HRI systems present a substantial challenge for software verification --- the process used to gain confidence in the correctness of an implementation, i.e.\ the robot's code, with respect to the requirements. {\em The robot responds to an environment that is multifaceted and highly unpredictable.} This is especially true for robots involved in direct interaction with humans, whether this is in an unstructured home environment or in the more structured setting of collaborative manufacturing. We require a verification methodology that is sufficiently realistic (models the system with sufficient detail) while thoroughly exploring the range of possible outcomes, without exceeding resource constraints. 
s. 
Prior work~\cite{BFS09:HRIshort,Cowley2011,Kouskoulas2013,Mohammed2010,Muradore2011,webster14formalshort} has explored the use of formal methods to verify HRI. Formal methods can achieve full coverage of a highly abstracted model of the interactions, but are limited in the level of detail that can practically be modelled. {\em Sensors, motion and actuation in a continuous world present a challenge for models and requirement formulation in formal verification.} Physical experiments or simulation-based testing may be used to achieve greater realism, and to allow a larger set of requirements to be verified over the real robot's code.  However, neither of these can be performed exhaustively in practice. 

{\em Robotic code is typically characterised by a high level of concurrency between the communicating modules} (e.g., nodes and topics used in the Robot Operating System, ROS\footnote{http://www.ros.org/}) that control and monitor the robot’s sensors and actuators, and its decision making. Parallels can be drawn here to the design of microelectronics hardware, which consists of many interacting functional blocks, all active at the same time. Hence it is natural to ask: `Can techniques from the microelectronics field be employed to achieve comprehensive verification of HRI systems?' 

In this paper, we present the use of Coverage-Driven Verification (CDV) for the high-level control code of robotic assistants, in simulation-based testing. CDV is widely used in functional verification of hardware designs, and its adoption in the HRI domain is an innovative response to the challenge of verifying code for robotic assistants. CDV is a systematic approach that promotes achieving coverage closure efficiently, i.e.\ generation of effective tests to explore a System Under Test (SUT), efficient coverage data collection, and consequently efficient verification of the SUT with respect to the requirements. The resulting efficiency is critical in our application, given the challenge of achieving comprehensive verification with limited resources. 

{\em The extension of CDV to HRI requires the development of practical tools that are compatible with established robotics tools and methods.} The microelectronics industry benefits from the availability of hardware description languages, which streamline the application of systematic V\&V techniques. No practical verification tool exists for Python or C++, common languages for robotics code~\cite{Trojanek2014}. A novel contribution of this paper is the development of a CDV testbench specifically for HRI; this implementation makes use of established open-source tools where possible, while custom tools have been created as necessary to complete and connect the testbench components (Test Generator, Driver, Checker and Coverage Collector). Additionally, we outline the relevant background to ensure robust implementation of CDV.

To demonstrate the feasibility and potential benefits of the method, we applied CDV to an object-handover task, a critical component of a cooperative manufacture scenario, implemented as a ROS and Gazebo\footnote{http://gazebosim.org/} based simulator. Our automated testbench conveniently allows the actual robot code to be used in the simulation. Model-based and constrained pseudorandom test generation strategies form the Test Generator. A Driver applies the tests to the simulation components. The Checker comprises assertion monitors, collecting requirement coverage. The Coverage Collector, besides requirement, includes code coverage. 

We verified selected safety and liveness (functional) requirements of the handover task to showcase the potential of CDV in the HRI domain.

The paper proceeds with an overview of the CDV testbench components and verification methodology in Section~\ref{sc:CDV}. The handover scenario is introduced in Section~\ref{sc:implementation}, where we then present the CDV testbench we used to verify the code that implements the robot's part of the handover task. Section~\ref{sc:results} discusses the verification and coverage results for this example. Conclusions and future work are given in Section~\ref{sc:conclusions}.
 
\section{Coverage-Driven Verification} \label{sc:CDV}

\subsection{Structure of a CDV Testbench}
In CDV, a verification plan must be constructed before the testing process begins~\cite{Piziali2004}. This plan includes the aspects of the SUT that need to be verified, e.g.\ a requirements list or a functional description of the SUT, and a coverage strategy. The coverage strategy indicates how to achieve effective coverage, i.e.\ the exploration of the SUT and advancement of the verification progress, through the design of the testbench components, especially the Test Generator, the Checker and the Coverage Collector. The coverage strategy also specifies how to measure the coverage, e.g.\ a requirements model or a functional model to traverse.

In Testing, the SUT is placed into a test environment, a {\em testbench}. The testbench represents (a model of) the universe, or of its target environment. The process of testing is realised using the following four core components in a testbench, as shown in Fig.~\ref{testbench}:

\begin{figure}[!t]
\centering
\includegraphics[width=0.8\textwidth]{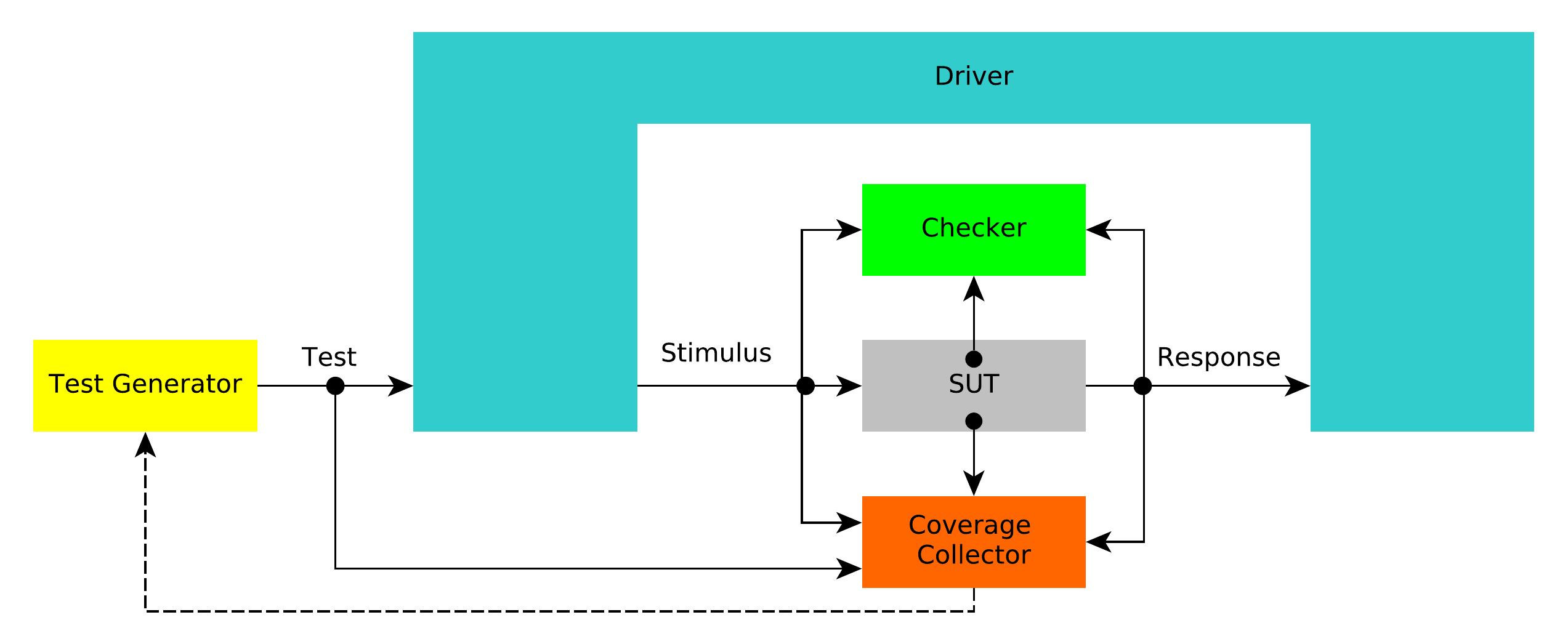}
\caption{Structure of a basic CDV testbench}
\label{testbench}
\end{figure} 

\begin{itemize}
\item the {\bf Test Generator} is the component that generates stimulus for
  the SUT;
\item the {\bf Driver} is the component that takes a test, potentially at a
  high level of abstraction, translates it into the level of abstraction used
  in the simulation, and drives it to stimulate the SUT;
\item the {\bf Checker} is the component that checks the response of the SUT to
  the stimulus and detects failures; 
\item the {\bf Coverage Collector} is the component that records the quality
  of the generated tests with respect to a set of complementing coverage models.
\end{itemize}

A key objective in the design of a CDV testbench is to achieve a fully autonomous environment, so that verification engineers can
concentrate on areas requiring intelligent input, namely efficient and effective test generation, bug detection, reliable tracking of progress and timely completion.

In the following sections we describe each testbench component in more detail, explaining how they can be used for verification in robotics.

\subsection{Test Generator}\label{sc:testgen}

The test generator aims to exercise the SUT for verification (activation of faults), while working towards full coverage. Test generators in CDV make use of pseudorandom generation techniques. Using pseudorandom as opposed to random generation allows repeatability of the tests. The generated tests must be valid (realistic, like a sensor input that reflects a valid scene). An effective set of tests includes a good variety that explores unexpected conditions and addresses the scenarios of interest as per the requirements list. An efficient set of tests maximises the coverage and verification progress, whilst minimizing the number of tests needed. To achieve the former while allowing for the latter, pseudorandom test generation can be biased using constraints.  These constraints can be derived from the SUT's functional requirements or from the verification plan~\cite{Piziali2004}. However, supplying effective constraints requires significant engineering skill and application knowledge. It is particularly difficult to generate meaningful sequences of actions, whether these are transactions on the interface of a system-on-chip, or interactions between humans and robots.
 
Constrained pseudorandom test generation can be complemented with model-based techniques~\cite{Haedicke2012,Lakhotia2009} to generate sequences that address specific use cases, such as interaction protocols between human and robot in a collaborative manufacturing environment. In model-based test generation, a model is explored or traversed to obtain abstract tests, i.e.\ tests at the same level of abstraction as the model. These abstract tests can serve as test templates, or constraints, for tests that target specific behaviours~\cite{Lackner2012,Nielsen2003}. For this, a model needs to be implemented, e.g.\ one that captures the intended behaviours of the robot when interacting with humans and/or its environment. In robotics, the degree of abstraction between such a model and the simulation often differs significantly compared to that observed in microelectronics~\cite{Nielsen2014}. Many low-level implementation details such as motion control, sensing models or path planning are abstracted from (e.g., as in~\cite{webster14formalshort}) to keep these models within manageable size.  For model-based testing to be credible and effective, the correctness of the behavioural model with respect to the robot's code needs to be established. However, this is beyond the scope of this paper.

\subsection{Driver}\label{sc:driver}

The Driver is a fully automated component that translates a (potentially high-level) description of a test into signal-level stimulus that can be applied to the interfaces of the SUT in order to expose the SUT to the situation prescribed by the test. The Driver may comprise an interacting network of modules corresponding to the distinct interfaces of the SUT. The SUT reacts to the stimuli provided on its interfaces. The Driver runs in parallel with the SUT and responds to it, if necessary; i.e., the Driver can be reactive. The automation of the Driver makes it feasible to execute batches of abstract tests, to accelerate testing.

In HRI, the Driver comprises a model of the human, a physics model, and communication channels to represent any interactions that do not require detailed physical simulation. For example, if the human element in the simulator is driving the robot's code, the Driver would execute the corresponding high-level action sequence, one item at a time, by translating it into the respective sequence of input signals, potentially passing through the physics model before exposing the signals to the robot's input channels.
 
\subsection{Checker}\label{sc:checker}
The automation of test generation prompts the need for automatic and test independent checkers, i.e.\ self-checking testbenches. Assertion-based verification~\cite{abd} allows locating checkers, in the form of assertion monitors, close to the code that is being observed.

Requirements to verify can be expressed as Temporal Logic properties. Assertion monitors can be derived automatically from these
properties~\cite{Havelund2002}, in an automata-based form. Since the simulations are time bound, some safety properties defined over infinite traces (e.g., using an \verb+always+ Temporal Logic operator) are bound over the duration of a simulation run. Relevant work in~\cite{Armoni2006} mentions the advantages of the automatic generation of monitors as automata, including the reduction of errors caused by manual translation.

For requirements about the low-level continuous behaviour of the SUT (e.g., trajectories computed by the motion planning), the monitoring can be performed in a quasi-continuous manner, considering computational limitations. Otherwise, over-approximations or interpolation can be performed to predict events at instants of time between computations, such as the overlapping of regions in the 3D space for collision avoidance.

\subsection{Coverage Collector}\label{sc:coverage}
Automatic test generation necessitates monitoring the quality of the tests to gain an understanding of the verification progress. To achieve this, statistics can be collected on the tests, the driven stimulus (external events), the SUT's response, and the SUT's internal state, including assertion monitors. In general, we distinguish between {\em code} coverage models and {\em functional} coverage models. A comprehensive account on coverage can be found in~\cite{Piziali2004}. 

The collected coverage data provides information on unexplored (coverage ``holes'') or lightly covered areas. {\em Coverage closure} is the process of identifying coverage holes and creating new tests that address these holes. This introduces a feedback loop from coverage collection/analysis to test generation, termed Coverage Directed test Generation (CDG)~\cite{Piziali2004}. Attempts have been made to automate CDG using machine learning techniques~\cite{ioannides}. However, CDG remains a difficult challenge in practice.

In principle, coverage collection and analysis techniques can be transferred directly into the domain of robotics verification. In fact, it is interesting to note that functional coverage in the form of ``cross-product'' coverage~\cite{wile}, as widely used in hardware design verification, has recently been proposed (independently) for the verification of autonomous robots in~\cite{Alexander2015}, where it is termed {\em situation} coverage and includes combinations of external events only.
 
\subsection{CDV Methodology}\label{sc:cdvmethod}
In CDV, an iterative process of test generation, execution, coverage collection and analysis is used to achieve coverage closure over several cycles. In practice, engineering input is required to interpret the data and to guide test generation towards closing coverage holes. This is either achieved simply by allowing further pseudorandom tests to be generated, by adding constraints to bias test generation, by employing model-based test generation or, as a last resort, by directed testing. If model-based test generation has already been applied, modifications to the formal model may yield new tests.  

It is important to note that further test generation is not always the only appropriate response to a coverage hole or a requirement violation.  The following options should also be considered: 1) the SUT has a bug, to be referred to the design team; 2) modifications to one or more of the requirements models (e.g.\ assertions or formal properties) are needed to more accurately reflect the actual requirements and/or design of the SUT; and/or 3) modifications to one or more of the testbench components are needed. This third decision may be reached if the tests and requirements models are deemed appropriate but the testbench does not allow the SUT's full range of functions to be exercised and observed.

\section{CDV Implementation}\label{sc:implementation}
A case study from a collaborative manufacture scenario is presented. We demonstrate the transferability of CDV into the HRI domain by constructing a CDV testbench for this case study using a combination of established open-source tools and custom components. Our implementation showcases the potential of CDV to verify robotic code used in HRI.

\subsection{Case Study: Robot to Human Object Handover Task}\label{Example}
Our case study is an object handover, a critical subtask in a broader scenario where a robot (BERT2~\cite{lenz2010bert2}) and a person work together to assemble a table. The handover starts with an activation signal from the person to the robot. The robot then picks up an object, and holds it near the person. The robot indicates it is ready for the person to receive the object. Then, the person is expected to hold the object simultaneously, moving closer if necessary, and to look at it --- indicating readiness of the person. The robot collects data through two different sensing systems: ``pressure'', sensors that determine whether just the robot, or simultaneously the robot and the person, are holding the object; and ``location'' and ``gaze'' sensors, an `EgoSphere' system that tracks whether the human hand is close to the object and whether the human head is directed towards the object~\cite{lenz2010bert2}. Based on the sensors, the robot determines whether the release condition is satisfied, and decides on a course of action: the robot will release the object and allow the person to take it, if the human was ready; if not, the robot will not release the object. The robot or human may disengage from the task (look or move away). The sensors are considered perfect. 

According to the handover task's interaction protocol, a robot ROS `node' was developed in Python, comprising 209 code statements. This node was structured as a state machine, using the SMACH modules~\cite{SMACH}, to facilitate modularity. 
The states, with their transitions, can be enumerated as shown below.  Each state transitions to the next in sequence, except where indicated otherwise. The code is also depicted as a flow chart in Figure~\ref{codecoverage}. 

\begin{enumerate}
\item \verb|reset| - The robot moves to its starting position, with gripper open.
\item \verb|receive_signal| - Read signals.  If `startRobot' is received, transition to \verb|move|; elseif timeout, transition to \verb|done|; else, loop back to present state.
\item \verb|move| - Plan trajectory of hand to piece. Move arm. Close gripper. Plan trajectory of hand to human. Move arm.
\item \verb|send_signal| - Send signal to inform human of handover start.
\item \verb|receive_signal| - Read signals.  If `humanIsReady' is received, transition to \verb|sense|; elseif timeout, transition to \verb|done|; else loop back to present state.
\item \verb|sense| - Read sensors. If timeout, transition to \verb|done|; elseif not all signals available, loop back to present state; else, transition to \verb|decide|.
\item \verb|decide| - If all sensors are satisfied, transition to \verb|release|; else, transition to \verb|done| (without releasing).
\item \verb|release| - Open the gripper.  Wait for 2 seconds.
\item \verb|done| - End of sequence.
\end{enumerate}

\subsection{Requirements}\label{ssc:requirements}
Requirements were derived from ISO~13482:2014 and desired functionality of the robot in the interaction~\cite{Grigore2011}: 

\begin{enumerate}
\item If the gaze, pressure and location are sensed as correct, then the object shall be released. 
\item If the gaze, pressure or location are sensed as incorrect, then the object shall not be released. 
\item The robot shall make a decision before a threshold of time. 
\item The robot shall always either time out, decide to release the object, or decide not to release the object.
\item The robot shall not close the gripper when the human is too close.
\end{enumerate}

Requirements 1 to 4 refer to sequences of high-level events over time, whereas Requirement 5 refers to a lower-level safety requirement of the continuous state space of the robot in the HRI. Thus, the former can be both targeted with model-based techniques and implemented as assertion monitors, whereas the latter is only suitable for implementation as an assertion monitor. 

\subsection{CDV Testbench Implementation}
\begin{figure}[h]
\centering
      \includegraphics[width=0.4\textwidth]{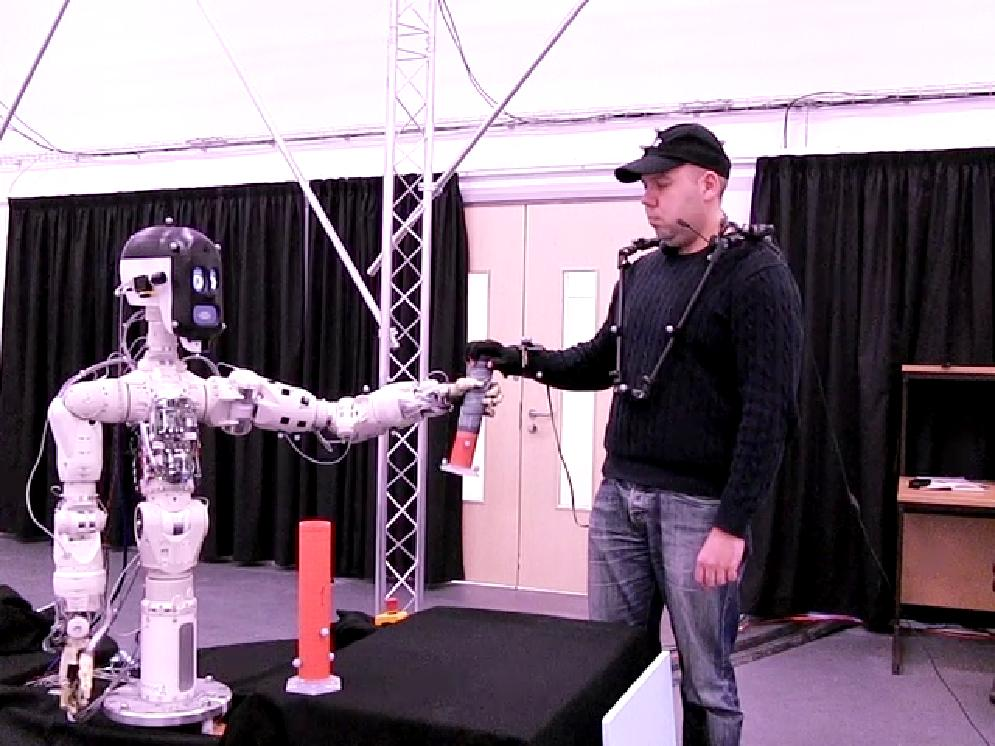}
      \includegraphics[width=0.4\textwidth, trim= 70mm 40mm 170mm 30mm, clip]{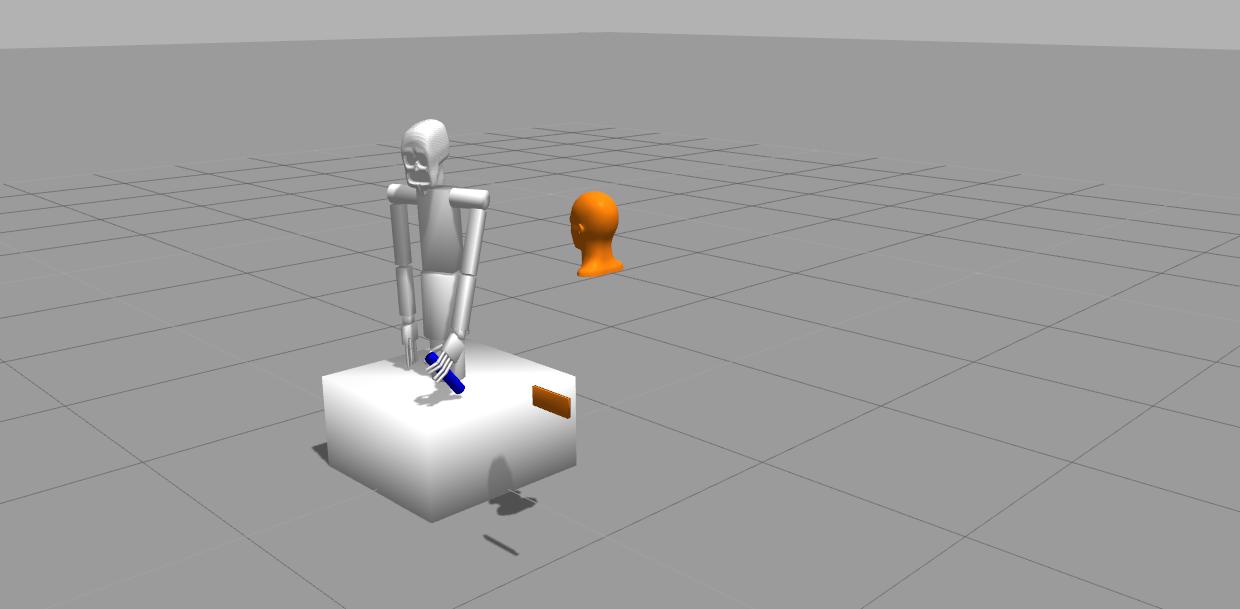}
\caption{BERT2 robot and a human, and the simulator in ROS-Gazebo}
\label{Simulatorphoto}
\end{figure}

ROS is a widely used open-source platform for the design of code for robots in C++ and/or Python. ROS allows interfacing directly with robots. Gazebo is a robot simulation tool designed for compatibility with ROS, that is able to emulate the physics of our world. Thus, the combination ROS-Gazebo provides a means of developing a robotic simulator, as shown in Figure~\ref{Simulatorphoto}. 

\begin{figure}[h]
\centering
\includegraphics[width=0.8\textwidth]{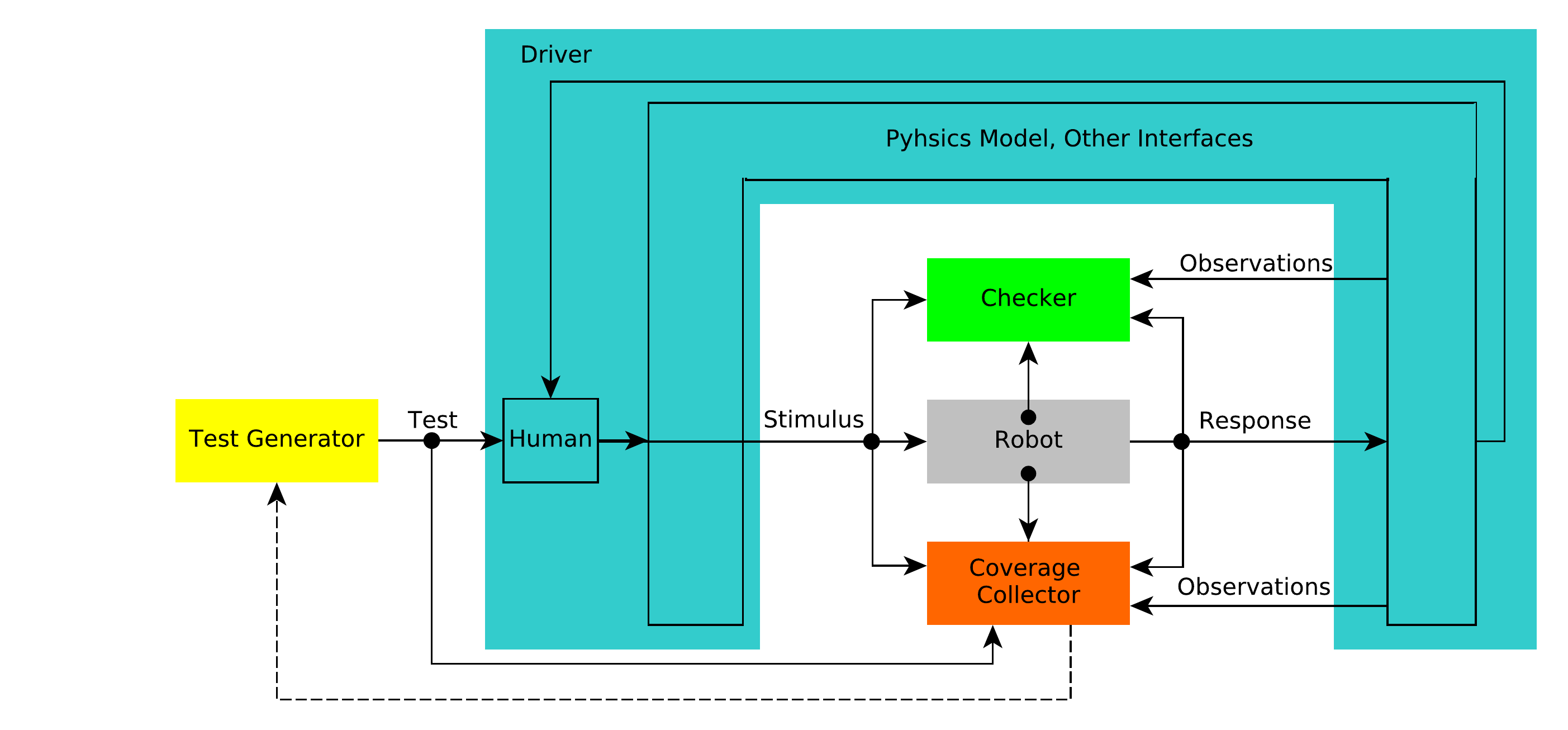}
\caption{Testbench and simulator elements in ROS-Gazebo}
\label{simstructure}
\end{figure}

Figure~\ref{simstructure} shows the structure of our CDV testbench implementation, incorporating the robot's high-level control code. The Driver incorporates the Gazebo physics simulator and the MoveIt!\footnote{http://moveit.ros.org/} packages for path-planning and inverse kinematics of the robot's motion. The human is embodied as a floating head and hand for simplicity; in future, this representation can be replaced by one that is anatomically accurate. The implementation in ROS ensures that assertion monitors  and coverage collection can access parameters internal to the robot code as well as the external physics model and other interfaces, such as signals. Observability of the external behaviour allows validating the robot's actions. In real life experiments, this is equivalent to observing the robot's physical behaviour to see if its responses are as expected.

\subsection{Test Generator and Driver}
Tests were generated pseudorandomly, by concatenating randomly selected elements from the set of high-level actions belonging to the handover workflow, forming random action sequences and instantiating relevant parameters. These randomized sequences represent environmental settings that do not necessarily comply with the interaction protocol. Thus, pseudorandom action sequence generation produces stimulus that correspond to unexpected behaviours that were not previously considered in the requirements. Posteriorly, constraints were introduced to bias the pseudorandom generation to obtain tests that do comply with the interaction protocol (e.g., enforcing particular sequences of actions). 

The handover interaction protocol was formalized as a set of six automata, in particular Probabilistic-Timed Automata (PTA) ~\cite{Hartmanns2009}, comprising the robot, the workflow, the gaze, the location, the pressure, and the sensors. Behaviours of the different elements (e.g., protocol compliant actions to activate the robot through signals) were abstracted in terms of state transitions and variable assignments. The structure of the robot's code guided the abstraction process, and the abstraction was verified via bisimulation analysis~\cite{CCSbook}.

Model-based test templates were obtained from witness traces (examples or counterexamples) produced by model checking the product automaton~\cite{Nielsen2003}. These witnesses contain combined sequences of states from the different automata. Requirements 1 to 4 (Section~\ref{ssc:requirements}) were used to derive model-based test templates that would trigger corresponding assertion monitors. We employed UPPAAL\footnote{http://www.uppaal.org/}, a model checker for PTA that produces witnesses automatically. Projections over these traces with respect to the workflow, gaze, location, pressure and sensors automata remove the elements that correspond to the robot's activities, to form a test template. Based on these test templates, tests were generated pseudorandomly. 

A test template for our simulator consists of a sequence of high-level actions (workflow) to activate the robot expressed as a state machine. A test comprises, besides the high-level actions, the pseudorandom instantiation of parameters, from well defined sets (e.g., ranges of values for gaze correct or gaze incorrect). An example is shown in Figure~\ref{exampletest}. The Driver produces responses in the physical model in Gazebo, signals to be communicated to the robot, and sensor readings. 
 
\begin{figure}[h]
\scriptsize
\centering
\begin{tabular}{r|lllll|lll}
\cline{2-6}
1&&\verb+sendsignal+&&\verb+activateRobot+ &&& &\\
2&&\verb+setparam+ && \verb+time = 40+& && This \verb+time+ instantiation produces &\\
3&&\verb+receivesignal+ && \verb+informHumanOfHandoverStart+ &&&a waiting time of $40 \times 0.05$ seconds.&\\
4&&\verb+sendsignal+ && \verb+humanIsReady+ &&&&\\
5&&\verb+setparam+ && \verb+time = 10+ &&&&\\
6&&\verb+setparam+ && \verb+honTask = true+ && &&\\
7&&\verb+setparam+ && \verb+hgazeOk = true+ && &Gaze instantiation for \verb+true+: choosing	offset,&\\
8&&\verb+setparam+ && \verb+hpressureOk = true+ && & distance and angle, from ranges $ \{[0.1,0.2],$&\\
9&&\verb+setparam+ && \verb+hlocationOk = true+ && & $[0.5,0.6],[15,40)\}$, e.g., $(0.1,0.5,30)$&\\ 
\cline{2-6}
\end{tabular}
\caption{Example test from a test template, comprising high-level actions and some parameter instantiations (time and gaze)}
\label{exampletest}
\end{figure}

An example of a constraint for constrained pseudorandom generation is the enforcement of the sequence of actions in lines 1 to 4 of Figure~\ref{exampletest}, followed by any other action sequence. This constraint ensures the immediate activation of the robot, when a simulation starts. 

An added benefit from the development of a formal model for test generation is that this allows formal verification through model checking~\cite{ClarkeMC}. Formal verification can thus complement CDV. However, properties that hold for abstract models must still be verified at the code level. Model checkers for code (e.g., CBMC\footnote{http://www.cprover.org/cbmc/}, Java PathFinder\footnote{http://javapathfinder.sourceforge.net/}) target runtime bugs in general, such as arrays out of bounds or unbounded loop executions. These are, however, at a different level than the complex functional behaviours we aim to verify. In~\cite{webster14formalshort}, the runtime detail is abstracted, giving way to high-level behaviour models where functional requirements can be verified with respect to the model only. 

\subsection{Checker}

Assertion monitors were implemented for all the requirements in Section~\ref{ssc:requirements}. Requirements 1 to 4 were translated first into CTL properties, and then automata-based assertion monitors were generated manually. This process will be automated in the future. For example, Requirement 1 corresponds to the property:
$$E <> sgazeOk \wedge spressureOk \wedge slocationOk \wedge releasedTrue.$$

The resulting monitor is triggered when reading a sensors signal indicating the gaze, pressure and location are correct. Then, the automaton transitions when receiving a signal of the object's release. If the latter signal event happens within a time threshold (3 seconds), a \verb+True+ result is reported. Finally, the automaton returns to the initial state.

Requirement 2 corresponds to the CTL property:
$$E<> (sgazeNotOk \vee spressureNotOk \vee slocationNotOk) \wedge releasedFalse.$$

This monitor is triggered when any of the gaze, pressure or location are incorrect in a sensing signal. Then, the automaton transitions to either a \verb+False+ result when receiving a signal of the object's release, or a \verb+True+ result if some time has elapsed (2 seconds) and no release signal has been received. Finally, the automaton returns to the initial state. 

Requirement 5 refers to physical space details abstracted from our PTA model, and it cannot be expressed as a Temporal Logic property. Hence, it was directly implemented as an automaton-based assertion monitor. When the robot grabs the object, it needs to make sure the human's hand (or any other body part) is at a distance. The monitor is triggered every time the code invokes the \verb+hand(close)+ function, which causes the motion of the robot's hand joints. The location of the human hand is then read from the Gazebo model (the head is ignored, since the model is abstracted to a head and a hand). If this location is close to the robot's hand (within a 0.05\,m distance of both mass centres), the monitor registers a \verb+False+ result, or otherwise \verb+True+. 

The monitors automatically generate report files, indicating their activation time, and the result of the checks if completed. 

\subsection{Coverage Collector}
We implemented two coverage models: code (statement) coverage and functional coverage in the form of requirements (assertion) coverage. The statement coverage was implemented through the `coverage'~\footnote{http://nedbatchelder.com/code/coverage/} module for Python. For each test run, statistics on the number of executed code statements are gathered. The assertion coverage is obtained by recording which assertion monitors are triggered by each test. If all the assertions are triggered at the end of the test runs, the testbench has achieved 100\% requirements coverage.

\section{Experiments and Verification Results}\label{sc:results} 
The CDV testbench described in Section~\ref{sc:implementation} was used 
\textit{(a)} to demonstrate the benefits of CDV in the context of HRI; 
\textit{(b)} to obtain an insight into the verification results, including unexpected behaviours or requirement violations; and 
\textit{(c)} to explore options to achieve coverage closure (from Section~\ref{sc:cdvmethod}).

The requirements mentioned in Section~\ref{Example} were verified using a CDV testbench in ROS (version Indigo) and Gazebo (2.2.5), and through model checking in UPPAAL (version 4.0.14), using the model we developed for model-based test generation. We used a PC with Intel i5-3230M 2.60\,GHz CPU, 8\,GB of RAM, running Ubuntu 14.04.

Table~\ref{results1} presents the assertion coverage for the handover, and the verification results from model checking. In model checking, the requirements were verified as true (T) or false (F). Through model checking, we were only able to cover Requirements 1 to 4. From each of the model checking witnesses (test templates) of Requirements 1 to 4, we generated a test (model-based generation). We also generated 100 pseudorandom (unconstrained) tests, and 100 constrained pseudorandom tests that enforced the activation of the robot as explained in Section~\ref{sc:implementation}. We verified Requirements 1 to 5 in simulation, and recorded the results of the assertion monitors: Pass (P), Fail (F), Not Triggered (NT), or Inconclusive (U) when the monitor was triggered but the check was not completed within the simulation run. The same setup was used to compute both assertion and statement coverage, allowing the comparison of the test generation strategies in terms of coverage efficiency. 

\begin{table}
\caption{Requirements (assertion) coverage and model checking results}
\centering
\scriptsize
\renewcommand{\arraystretch}{1.2}
\begin{tabular}{|c|c|cccc|cccc|cccc|}
\hline 
Req. & Model	& \multicolumn{12}{|c|}{Simulation-Based Testing} \\ \cline{3-14}
	& Checking&\multicolumn{4}{|c|}{Pseudorandom} & \multicolumn{4}{|c|}{Constrained-Pseudorandom} &\multicolumn{4}{|c|}{Model-Based}  \\
 \cline{3-14}
			&  				& P & F & NT & I		 & P & F & NT & I 			   & P & F & NT & I\\
\hline
1 		& T 	& 0/100  &0/100 & 100/100 & 0/100		& 0/100  & 0/100 & 100/100 & 0/100 	&  3/4 & 0/4 & 1/4 & 0/4\\
2 		& T 	& 33/100 & 0/100 & 67/100  & 0/100		& 87/100 & 0/100 & 13/100 & 0/100	&  1/4 & 0/4 & 3/4 & 0/4\\
3 		& T 	& 33/100 & 0	/100 & 67/100 & 0/100		& 87/100 & 0/100 & 13/100 & 0/100	&  4/4 & 0/4 & 0/4 & 0/4 \\
4 		& T 	& 98/100 & 0/100 & 0/100  & 2/100		& 98/100 & 0/100 & 0/100 & 2/100		&  4/4 & 0/4 & 0/4 & 0/4\\ \hline
5		& - 	& 46/100 & 0/100 & 54/100 & 0/100 		& 93/100 & 0/100 & 7/100 & 0/100 	&  4/4 & 0/4 & 0/4 & 0/4\\
\hline
\end{tabular} \label{results1}
\end{table}

The results in Table~\ref{results1} confirm our expectations for the different test generation strategies. For assertion-based functional coverage, pseudorandom and constrained-pseudorandom test generation are less efficient than model-based test generation, which triggered all five assertions with just four tests. Requirement 1 was not covered by either the pseudorandom or the constrained pseudorandom strategy. If either of these strategies was used alone, the coverage hole could potentially be closed by adding further constraints or by using a more sophisticated test generation strategy such as model-based test generation.

The assertion monitor checks for Requirement 4 were inconclusive for some of the pseudorandom and constrained-pseudorandom generated tests. This occurs because in these tests the robot is activated long after the start of the handover task (when the robot is reset and proceeds to wait for a signal). These tests do not comply with the protocol which requires to activate the robot at the start and within a given time threshold. 

This coverage result could trigger different actions, e.g.\ the assertion monitor could be modified to choose either pass or fail at the end of the simulation; the Driver could be modified such that the simulation duration is extended; or, the inconclusive checks could be dismissed as trivial, in which case the efficiency of any further tests could be improved by directing them away from such cases. As noted in Section~\ref{sc:cdvmethod}, further test generation is not always the sole appropriate response to a coverage hole. It is worth noting that this scenario was exposed only by pseudorandom and constrained-pseudorandom test generation, demonstrating the unique benefit of these approaches; by exploring the SUT's behaviour beyond the assumptions of the verification engineer, they provide a useful complement to the more directed approach of model-based test generation. 

\begin{figure}[h]
 \subfloat[\label{subfig-3:a}]{
      \includegraphics[width=0.27\textwidth]{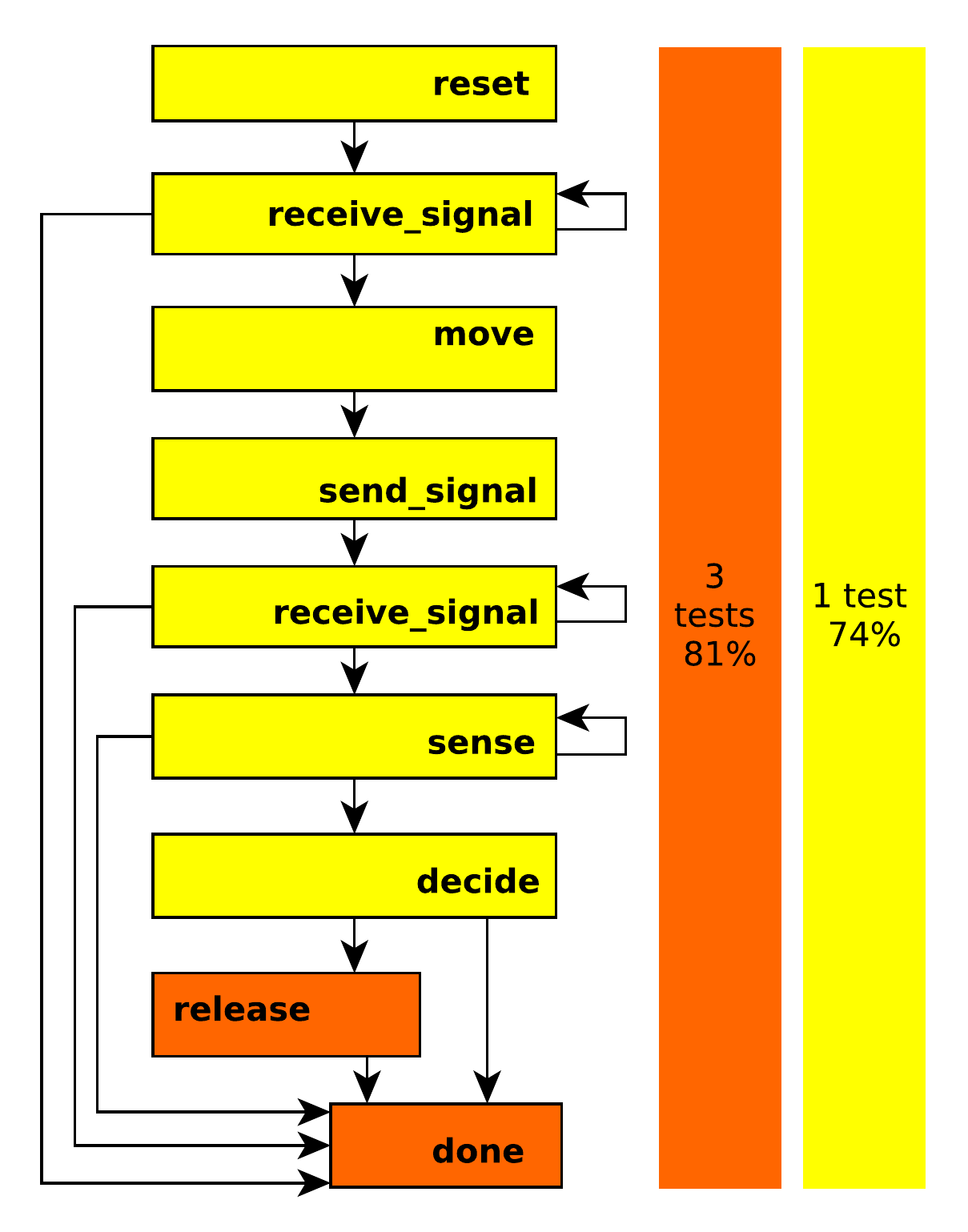}
    }
    \hfill
    \subfloat[\label{subfig-3:b}]{
      \includegraphics[width=0.36\textwidth]{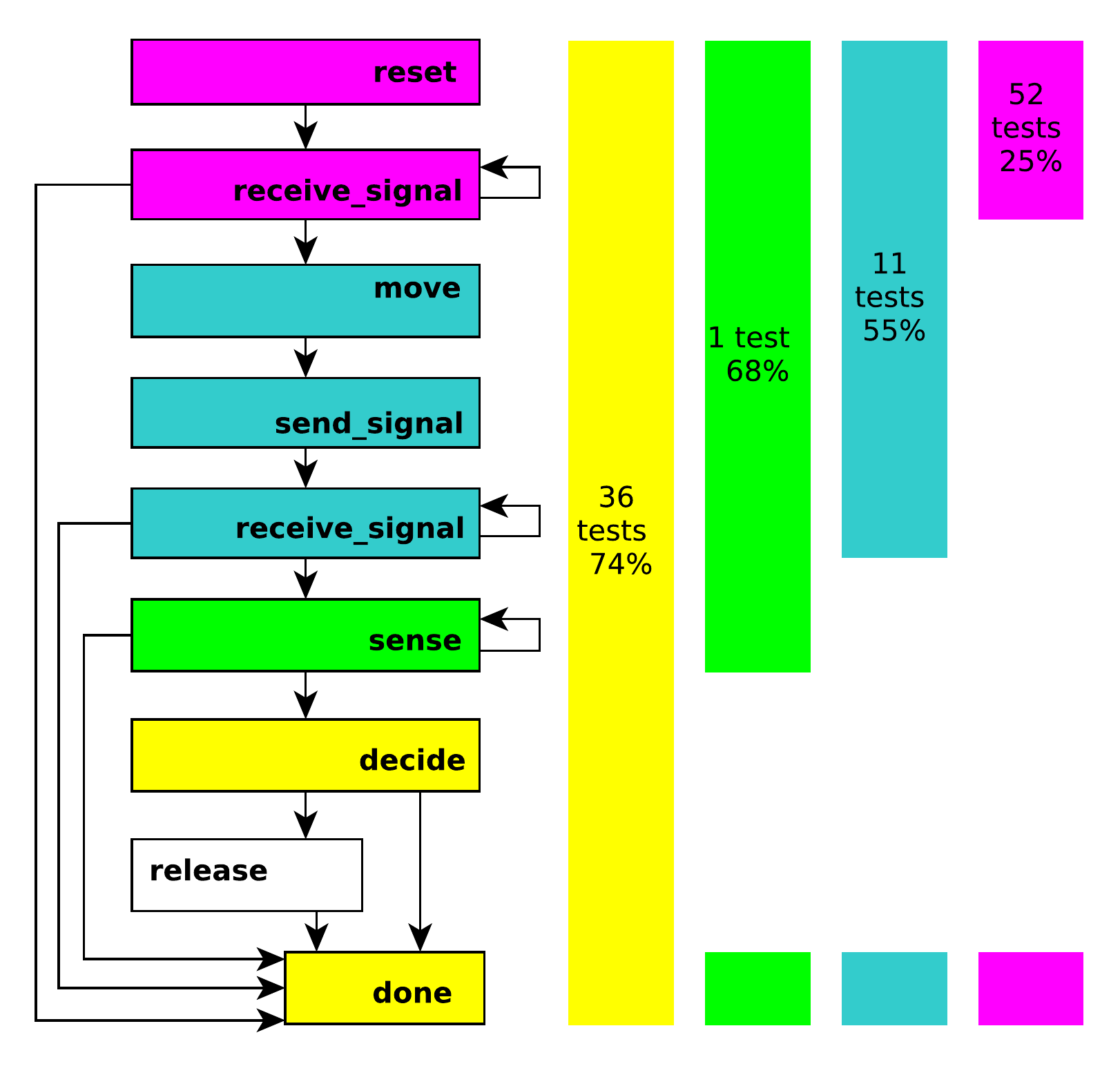}
    } 
    \hfill
    \subfloat[\label{subfig-3:c}]{
      \includegraphics[width=0.27\textwidth]{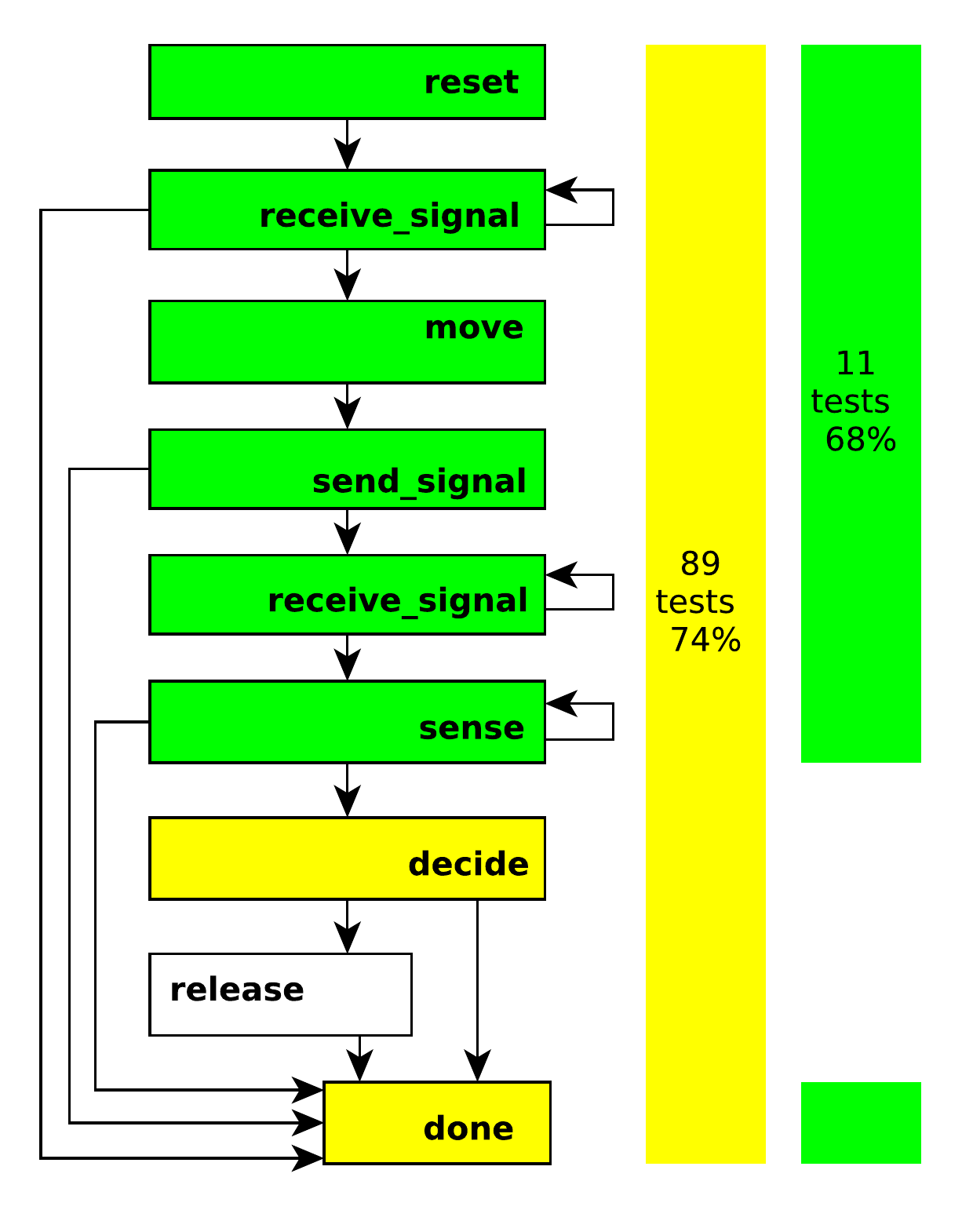}
    }
\caption{Code coverage (percent values) obtained in simulation with (a) model-based (4 tests), (b) pseudorandom (100 tests), and (c) constrained-pseudorandom test generation (100 tests)
}
\label{codecoverage}
\end{figure}

Figure~\ref{codecoverage} illustrates the code coverage (statements) achieved with each test generation strategy over 206 statements (the actual percentages may vary $\pm 2$\% due to decision branches with 1 or 2 lines of code each). The lines of code are grouped using the state machine structure in the Python module, to facilitate visualization. The block of code corresponding to the ``release'' state is not covered by the pseudorandom and constrained pseudorandom generated tests, hence it is shown in white. This coverage hole could be closed by applying the test template produced by model-based test generation for Requirement 1. 

Because our code is structured as a finite state machine (FSM), it would be appropriate to also incorporate structural coverage models in the future.  A comprehensive test suite would include tests that visit all states, trigger all possible state transitions, and traverse all paths.

The generation of effective tests, that target both the exploration of the SUT and the verification progress, is fundamental to maximising the efficiency of a CDV testbench reaching for coverage closure. From the overall results, it can be seen that the three test generation approaches applied have complementary strengths that overcome their respective weaknesses in terms of coverage. While model-based test generation ensures that the requirements are covered in an efficient manner, pseudorandom test generation can construct scenarios that the verification engineer has not foreseen. Such cases are useful for exposing flawed or missing assumptions in the design of the testbench or the requirements.

\section{Conclusions}\label{sc:conclusions}

We advocated the use of CDV for robot code in the context of HRI. By promoting automation, CDV can provide a faster route to coverage closure, compared with manually directed testing. CDV is typically used in Software-in-the-Loop simulations, but it can also be used in conjunction with Hardware-in-the-Loop simulation, Human-in-the-Loop simulation or with emulation.
The flexibility of CDV with regard to the levels of abstraction used in both the requirements models and the SUT makes it particularly well suited to verification of HRI.

The principal drawback of CDV, compared with directed testing, is the overhead effort associated with building an automated testbench. Directed testing produces early results, but CDV significantly accelerates the approach towards coverage closure once the testbench is in place. Hence CDV is an appropriate choice for systems in which complete coverage is difficult to achieve due to a broad and varied state space that includes rare but important events, as is typically the case for HRI.

We proposed implementations of four automatic testbench components, the Test Generator, the Driver, the Checker and the Coverage Collector, that suit the HRI domain. Different test generation strategies were considered: pseudorandom, constrained pseudorandom and model-based to complement each other in the quest for meaningful tests and exploration of unexpected behaviours. Assertions were proposed for the Checker, accommodating requirements at different levels of abstraction, an important feature for HRI. Different coverage models were proposed for the Coverage Collector: requirements (assertion), code statements, and cross-product. 

The potential for CDG (Coverage-Driven test Generation), through the implementation of automated feedback loops, has been considered. Nevertheless, we believe a great part of the feedback work needs to be performed by the verification engineer, since CDG is difficult to implement in practice. 

A handover example demonstrated the feasibility of implementing the CDV testbench as a ROS-Gazebo based simulator. The results show the relative merits of our proposed testbench components, and indicate how feedback loops in the testbench can be explored to seek coverage closure. Several key observations can be noted from these results. Pseudorandom test generation allows a degree of unpredictability in the environment, so that unexpected behaviours of the SUT may be exposed. Model-based test generation usefully complements this technique by systematically directing tests according to the requirements of the SUT. This requires the development of a formal model of the system, which additionally enables exhaustive verification through formal methods, as explored by previous authors for HRI~\cite{BFS09:HRIshort,Cowley2011,Kouskoulas2013,Mohammed2010,Muradore2011,webster14formalshort}.

If the requirements are translated into Temporal Logic properties for model checking, assertion monitors can be derived automatically. In future work, we will be exploring generation of monitors for different levels of abstraction in the simulation (e.g., events-based, or checked at every clock cycle) in a more formal manner. We will further explore the use of bisimulation analysis to ensure equivalence between a robot's high-level control code and any associated formal models. We intend to incorporate probabilistic models of the human, the environment and other elements in the simulator, to enable more varied stimulation of an SUT. We also intend to verify a more comprehensive set of requirements for the handover task, e.g., according to the safety standard ISO~15066 (currently under development) for collaborative industrial robots.

Our approach is scalable, as more complex systems can be verified using the same CDV approach, for the actual system's code. We are confident CDV can be used for the verification and validation of autonomous systems in general. Open source platforms and established tools can serve to create simulators and models at different abstraction levels for the same SUT.  

\paragraph{\bf Acknowledgments:}

This work was supported by the EPSRC grants\\ EP/K006320/1 and EP/K006223/1 ``Trustworthy Robotic Assistants''. 

We are grateful for the productive discussions with Yoav Hollander, Yaron Kashai, Ziv Binyamini and Mike Bartley.

\bibliographystyle{plain}

\bibliography{references}

\end{document}